# Perception, Attention, and Resources: A Decision-Theoretic Approach to Graphics Rendering

Eric Horvitz and Jed Lengyel

Microsoft Research
Redmond, Washington 98052
{horvitz, jedl}@microsoft.com

## Abstract

We describe work to control graphics rendering under limited computational resources by taking a decision-theoretic perspective on perceptual costs and computational savings of approximations. The work extends earlier work on the control of rendering by introducing methods and models for computing the expected cost associated with degradations of scene components. The expected cost is computed by considering the perceptual cost of degradations and a probability distribution over the attentional focus of viewers. We review the critical literature describing findings on visual search and attention, discuss the implications of the findings, and introduce models of expected perceptual cost. Finally, we discuss policies that harness information about the expected cost of scene components.

## 1 Introduction

The rendering of high-quality 3D graphics is computationally intensive. We describe the use of decision-theoretic models and methods to control rendering approximations under limited or varying computational resources. After discussing several dimensions of degradation associated with rendering approximation strategies, we review key results on visual search and attention elucidated by cognitive psychologists. The results provide context and guidance for building models of visual attention and of the expected cost of graphics approximations. Following a discussion of the implications of the psychological results, we summarize our ongoing perceptual studies of image quality, and present a set of models that show promise for guiding the control of rendering approximation. Finally, we discuss how the models can be used to generate policies for rendering scenes. We describe application of the methods within the Talisman graphics architecture. The Talisman project at Microsoft [32] has focused on the definition of a flexible multimedia architecture with the ability to deliver high-quality graphics, audio, and video on personal computers.

In an earlier paper, we surveyed key problems and opportunities with the decision-theoretic control of multiple dimensions of rendering in systems based on a layered graphics pipeline [13]. In this paper, we present additional details on building and using perceptual models for regulating rendering under scarce resources. Our overall concerns and goals with tradeoffs in graphics are related to the work of Funkhouser and Sequin [9], who have explored adaptive strategies for selectively controlling the level of detail of objects in the interactive visualization of architectural models. In distinction to prior work, we decompose the task of modeling the expected cost of scenes into (1) the construction of models of perceptual loss for image components, (2) the development of probabilistic models of attention that transform costs to expected costs, and the (3) assembly of costs associated with different components into a cost for an entire image or segment of images. We describe how measures of the expected cost of image components can be used in regulation systems that seek to minimize the expected perceptual loss of images given limited or varying resource constraints.

## 2 Dimensions of Rendering Approximation

There has been growing interest in the development of approximation strategies and architectures for rendering complex scenes. In several recent approaches, images are decomposed into sets of scene elements which are manipulated separately. We can decompose an image in a variety of ways, and can employ different approximation strategies to degrade the fidelity with which the individual components of scenes are rendered. In the Talisman test bed, images are decomposed into a set of *sprites* which can be subjected to individual approximations. *Sprites* are shaped image layers with an associated two-dimensional transforma-tion. Each pixel in the sprite image has, in addition to the color channels, an alpha channel which allows for non-rectangular image shapes and smoothly blended boundaries. Sprites are selected based on the stability in the structure and common trajectories of objects in an image. We shall use sprites broadly in our discussion to refer to a set of elements comprising an image and the underlying geometric models, textures, materials, etc., needed to render each element's image.

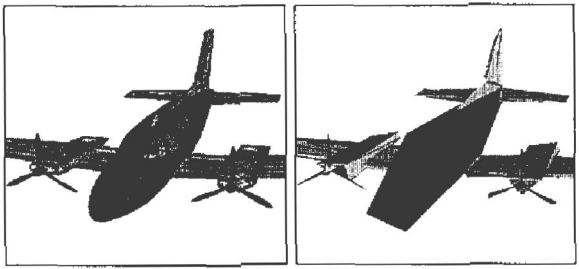

Figure 1: Strategies for reducing the complexity of models. Algorithms for reducing incrementally the number of vertices used to describe a geometric model can be used to generate a spectrum of models from an ideal model to progressively simpler models. The fully detailed model (left) is described by 6,795 vertices. The simpler model (right) is described by 97 vertices.

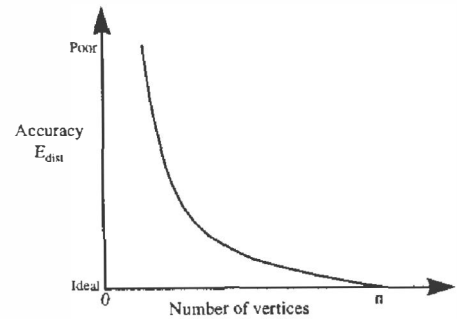

Figure 2: Tradeoff between size and accuracy for the progressive mesh method. The graph displays the relationship between a measure of model accuracy based on an energy quantity ($E_{dist}$), that summarizes the differences between an ideal model and simpler models produced by the progressive mesh approximation procedure.

Let us explore several graphics approximations strategies. We will focus in particular on flexible strategies, methods which can provide for the smooth degradation in the fidelity of sprites, along one or more dimensions of degradation, in return for diminishing computational requirements. Flexible rendering strategies include approximations that allow for the graceful reduction of model complexity, spatial resolution, shading complexity, and for the rate at which sprites are updated. Typically, we can generate, for each of the degradation dimensions, a corresponding measure of error, or fiducial, which estimates the distance from a gold-standard rendering along that dimension of quality.

The dynamic selection of simpler models from a spectrum of models to enhance rendering performance (referred to as *detail elision*) has been employed in a number of graphics systems. Model-simplification methods include the storage of multiple models, the dynamic simplification of the mesh of polygons that specify a model, and the choice of sampling for surfaces specified by parametric models. As the geometry used to describe a model is simplified, details of its structure are lost and artifacts appear in the form of errors in the overall shape or surface characteristics of objects.

An example of a flexible model-simplification strategy is the *progressive mesh* developed by Hoppe [12]. With this approach, edges of polygons composing a mesh are selected for collapse into single vertices so as to minimize a measure of energy defined in terms of multiple factors including the total number of vertices and the squared distance of points in the simplified models from an ideal mesh. The edge-collapse operation is reversible with tractable vertex-splitting operations, allowing for rapid traversal of a spectrum of meshes. Figure 1 displays a fully detailed model and a point on the spectrum of increasingly simplified models developed by the progressive mesh method. Figure 2 shows the relationship between the number of polygons included in a model and a measure of model accuracy based on a summary energy measure.

Spatial resolution methods center on the control of the number of samples dedicated to a sprite and by the image compression factor. We can selectively reduce the number of pixels devoted to a sprite. As the resolution is diminished, objects lose detail and ultimately become granular and fuzzy. In Talisman, the sprite transformation allows the sprite to be rendered with a low number of pixels and interpolated to the higher screen resolution. The quality of the resulting blur depends on the quality of the image filtering.

The shading complexity is determined by the type of texture filtering, texture level-of-detail, the number of lights used to illuminate a scene, and the use of shading effects such as reflections or shadows. Diminishing shading complexity introduces granularity and artifacts in the subtleties of lighting and reflection. For example, a reduction in the texture level-of-detail blurs the texture applied to the object. The magnitude of error can be captured by photometric estimates.

*Temporal resolution* methods center on the control of the rate at which scenes or components of scenes are updated. Temporal resolution methods include the simple approach of reducing the frame rate to allow enough computation to render scenes. This common approach leads to problems with flicker and is especially costly in interactive applications. A more sophisticated temporal-resolution strategy is to maintain frame refresh rate, but to render objects within frames at different frequencies depending on the configuration and dynamics of objects.

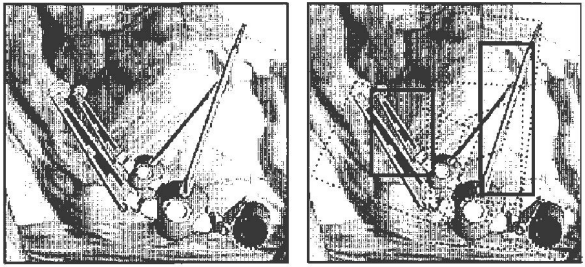

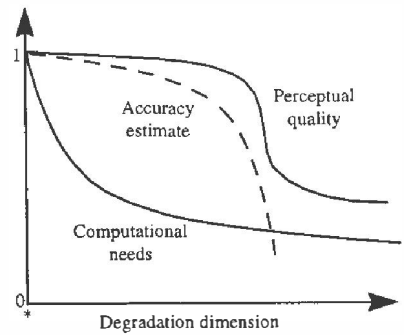

Figure 3: Minimizing computation through reuse of sprites. Talisman takes advantage of spatial and temporal coherence by attempting to transform previously rendered sprites rather than re-rendering the sprites. Rectangles composed of dotted lines bound sprites warped to fit a new frame; rectangles of solid lines bound re-rendered sprites.

Figure 4: Considering perceptual costs and computational savings. The schematized graphs highlight our pursuit of an understanding of relationships between measures of simplification, and perceptual and computational costs as image elements are degraded.

The manipulation of temporal resolution of individual sprites is central in Talisman. In Talisman, computationally inexpensive transformations are applied to update sprites generated from previous frames so as to approximate the general motion of an object or change of viewpoint in three dimensions—instead of undertaking more expensive re-rendering of the sprites from 3D models. Talisman attempts to minimize the computational requirements of rendering by employing inexpensive 2D transformations on previously rendered sprites rather than re-rendering sprites where possible. Where possible, computationally inexpensive 2D affine transformations are employed to update sprites generated for previous frames so as to approximate the general motion of an object or change of viewpoint in three dimensions—instead of undertaking more expensive re-rendering of the sprites from 3D models. Thus, Talisman attempts to adapt previously rendered sprites, bypassing the more expensive task of re-rendering the sprites, but potentially inducing spatial and temporal artifacts. Several groups have explored the similar use of inexpensive warping transformations on subcomponents of images to reduce realtime computation [25, 27, 3, 20].

In Talisman, sprites rendered in an earlier frame will continue to be transformed and reused in new frames until an estimate of error exceeds a tolerated error level. A small number of positions in the image, called characteristic points, are used to monitor and estimate the error as a function of the distances between the reused sprite and the actual sprite in the new frame. As the level of tolerated error is increased, the number of sprites that need to be re-rendered in each frame typically drops. However, as the error threshold is increased, geometric and temporal artifacts become more salient as sprites that have been warped for several frames are replaced with a re-rendered sprite after the larger error thresholds have been exceeded. By increasing the tolerated error, we save on resources by minimizing re-rendering, but may introduce noticeable artifacts in object distortion, problems with visibility of overlapping sprites, and discontinuities.

Figure 3 highlights the savings that can be achieved with sprite reuse. The left panel of the figure displays a frame drawn from a sequence of frames produced by a Talisman simulation that depicts two spacecraft traveling over mountainous terrain. The right panel shows a set of boxes that bound the individual sprites in the scene. The sprites that have been re-rendered for this frame are highlighted with solid lines, while the larger number of sprites that have been reused through inexpensive transformations are bounded with rectangles composed of broken lines.

## 3   Perception and Computation

Optimization of graphics rendering under varying resource constraints poses challenges at the interface between computation and cognition. We seek to model how degradations along different dimensions of rendering approximation can influence the perception of the quality of a scene. In addition, we wish to understand how a user's attention to different components of an image can change impressions of quality.

We may have access to information of the form portrayed in Figure 2, detailing the relationships between simple summaries of errors and alternate simplifications generated by flexible strategies. However, to understand the actual perceptual costs and benefits, we need to make the additional link to computational resources and perceptual accuracy, and to extend our understanding to multiple combinations of degradation. This goal is captured by Figure 4, which highlights the missing links between measures of accuracy and estimates of computational load, and between measures of accuracy and the perceived quality of images.

The overall aim of capturing the influence of multiple factors on the perceptual quality of images is the development of a rich, multiattribute perceptual utility model that represents the perceptual cost associated with the degradation of images from a gold-standard image or set of images. Attempts to map dimensions of image degradation to perception of quality brings us into the realm of the psychology of visual perception.

## 4  Findings in Visual Perception

The cognitive psychology of vision is an active area of research fraught with competing and complementary theories, numerous studies, and an array of interesting results. Several areas of investigation in visual perception have relevance to our pursuit of links between perceptual quality of images and rendering decisions.

As highlighted by the schematized functions displayed in Figure 5, we wish to move beyond information displayed in Figure 2 to map the links between perceptual quality and the metrics used to characterize the degradation of graphics along different dimensions (e.g., total squared distance deviation of mesh, tolerance of warp error-estimate, diminished resolution). We also need to characterize how various combinations of degradations, can influence subjective impressions of the quality of a scene–and to understand how attention to different aspects of a scene influences the cost of degradations. After all, our goal is to provide content that is visually satisfying to people. Simple scientific goals focused on maximizing precision across the board are likely to be naive from the point of view of genuinely optimizing the final visual result.

### 4.1  Visual Search and Attention

The paradigm in psychology for measuring the ability of human subjects to identify various features in scenes centers on a visual-search methodology. Studies of visual search have attempted to measure the abilities of human subjects to notice various components of scenes. A large set of studies have uncovered two interrelated classes of visual processing, referred to as preattentive and attentive vision, respectively. *Preattentive* vision is thought to continually scan large areas at a time in parallel, efficiently noting features representing basic changes in pattern or motion. *Attentive* visual processes refer to the more serial, resource-limited processes found to be required to recognize details about objects and relationships in scenes.

Neisser noted that features efficiently detected by the preattentive visual processes include the overall color, size, luminance, motion, temporal onset of patterns, and simple aspects of shape like orientation, curvature (but not closure, gaps or terminators) [21]. Julesz defined a class of features efficiently discriminated by preattentive vision, referred to as *textons* [17]. Textons include elongated shapes such as ellipses, rectangles, and line segments that have specific colors, widths, lengths, orientations, depths, velocities, and flicker. Textons also include the ends of line segments, referred to as terminators, and the crossing of lines. Preattentive processing has been shown to be limited in its ability to detect the absence of features in a target amidst a sea of similar targets that contain the feature (e.g., finding a circle without a slash among circles with a slash). More generally, the parallel, preattentive processes cannot efficiently identify cases where distinct features are conjoined into higher-level patterns or objects; identifying conjunctions of features requires the more focused, resource-strapped attentive vision processes.

Several studies have further elucidated the links between preattentive and attentive processes. For example, researchers have found that objects may be recognized rapidly through efficient interactions of preattentive and attentive processes and search can be made more efficient through training. An example of efficient recognition of objects is the "pop out" effect, where objects seem to jump out of background patterns. Wolfe, *et al.* performed studies suggesting that serial search for conjunctions can be guided and made more efficient taking advantage of parallel processes [34]. The group proposed that preattentive processes can filter out distracters from candidates, and, thus, reduce the size of the serial search. This effect appears to depend on the quality of the guidance provided by the parallel processes, and enhanced when elements are distinguished by luminance and color contrast, or when there are discontinuities in spatial or temporal patterning of the first-order properties giving rise to motion or texture differences [2]. Treisman and Gelade have studied the ability of people to recognize conjunctions of features. The team proposed and found evidence for the *feature-integration* theory of attention, where features are detected early on but are only related to one another to form recognizable objects with focused attention [33]. They also showed that recognition tasks were diminished by distraction and diverting of attention.

Investigation of visual attention has also explored the realm between preattentive and attentive processes by seeking out context-dependent changes in perceptual abilities. Under some conditions, visual resources appear to be distributed evenly over a display, with apparent parallel processing of display items [18, 29]. In other situations, a focused, serial scanning of items in a display occurs [7, 24]. One study showed that it is difficult for viewers to split their detailed visual attention to two separate spatial regions [23]. Eriksen and Hoffman have found the tight focus to have a diameter of approximately 1 degree of visual angle. More generally, response time and accuracy in search tasks have been found to be influenced by the spatial relationships between a target object, the position and number of related objects, and the location on the screen that subjects have been cued to focus on via verbal commands or visual stimuli [4, 31]. Reaction times in search tasks have been found to be fastest if attention is focused on the location of the target [23], and to increase as the visual angle between a current focus and target increases, as well as with the diminishing distance between a target and distractor objects [8, 11, 22, 30].

Capacity models for attention have been developed that attempt to describe the spatial characteristics of visual attention. Several different classes of model have been proposed including those where attentional resources are considered as being allocated simultaneously in different amounts to different locations in the visual field [28], models where resources are toggled between a distributed and a focused, "spotlight" state [16], and a "zoom-lens" model where there is a tradeoff between the breadth and degree of attention, and where resources are distributed evenly within the purview of the lens, from large views at low accuracies or power, to tighter focuses that are used to serially explore small areas with high-resolution [8, 7, 6, 15]. Several experiments have provided some evidence for the spotlight model [16] and for the zoom-lens model [8, 5, 1]. Nevertheless, there is still uncertainty about which models best describe visual attention, and to the extent that preattentive and attentive processes are distinct or can be explained by a larger model of attention.

### 4.2   Implications of the Findings

We wish to harness knowledge about the human visual system to model how viewers perceive degradations produced by rendering approximation. The basic findings in visual search provide intuitions that are useful for constructing parameterized models of perceptual loss. Such models can be refined by performing studies with subjects viewing a range of approximations produced by a graphics system.

In particular, results about the difficulty of viewers recognizing objects and missing details in objects in time-pressured search have implications in models of perceptual cost and of attention. The finding that object recognition is largely a function of attentive procedures and that attentive vision is a serial, resource-constrained process suggests that the perceptual cost of selective degradations is likely to be sensitive to the details of a viewer's attention, and that perceptual losses are minimized when the viewer is attending to portions of an image that are not significantly degraded.

The findings also characterize spatial and temporal features that guide a viewer's attentive processes to recognize objects and patterns more efficiently. Such results suggest that rendering artifacts may be especially prominent if they lead to discontinuities in such attributes as color, texture, or motion may lead to a "popping-out" of artifacts. The studies of attention can provide us with insights for developing models that relate the ease with which a viewer may notice objects as a function of their spatial relationships with an object at the center of attention.

To further extend our understanding of the links between the results from visual search research and the perceptual quality of images rendered by different regulation policies, we have been collaborating with cognitive psychologists on sets of perceptual studies aimed at directly exploring the influence of various rendering policies on the overall perception of image quality. We are investigating the perception of loss of quality as a function of dimensions of degradation provided in the Talisman system, including diminishment of spatial and temporal resolution individually and in combination. [14]. We have also been performing studies with gaze-tracking equipment to gain an understanding of attention as a function of content and to characterize perception of image quality as a function of attention.

## 5   Modeling the Cost of Rendering Approximation

The serial nature of attentive vision highlights opportunities for building models of degradation as a function of focus of attention, and for dynamically allocating resources based on attention. We shall now describe our work to build models for controlling graphics rendering inspired by the results on human visual perception and attention.

We decompose the modeling task for attention-based rendering into the tasks of (1) building a model of cost of perceptual degradation capturing the perceived losses as sprites are displayed with diminishing resources, (2) developing a model of attention, which provides probabilities that objects in a scene will be at the focus of a viewer's attention, and (3) providing a model that combines the costs of degradation associated with multiple sprites into the comprehensive cost of an entire frame or sequence of frames. Such models provide a foundation for refinement with empirical studies of cost and attention.

We shall consider first deterministic measures of perceptual cost. Then, we will develop a measure of expected perceptual cost based on a consideration of a probability distribution over foci of attention. In Section 6, we will discuss the authoring or generation of models of attention for providing the likelihoods that viewers will focus on particular aspects of a scene.

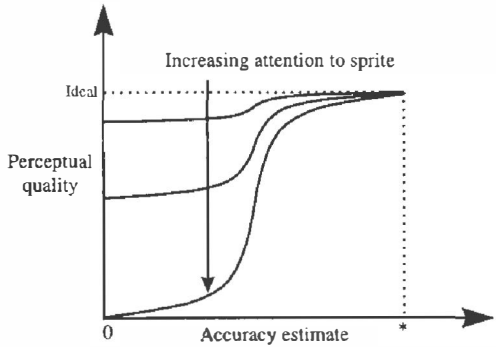

Figure 5: Attention and perception. These schematized graphs highlight our goal to better understand relationships between error metrics and perceptual cost, as well as the influence of attention on the perception of the quality of image elements.

## 5.1 Capturing Perceptual Cost

Let us first focus on the development of a perceptual cost function for individual sprites. A perceptual cost function, $C^p(R_k, S_i)$, provides us with a measure of cost associated with each sprite $S_i$ as a function of the rendering action $R$ performed on that sprite. This function captures the contribution of each sprite to the overall perceived quality of an image as a function of a measure of error between a gold standard rendering and an approximation induced by the rendering action. We shall assume that the perceptual loss associated with a perfect rendering of components is 0 and has a maximal cost that is a monotonically increasing function of (1) a measure of rendering error and (2) a measure of the visual salence of the image component. Visual salence may be associated with such variables as the total screen area occupied by a sprite. The details of the error measure and the visual salence of the component can be refined by specific visual perception studies.

For the general case, $R$ for each component represents a vector of decisions about the approximation of component rendering along the different dimensions of degradation. In Talisman, key dimensions of degradation include temporal approximation via re-use of sprites with an affine transformation. We have constructed cost models that employ a cost function computed as the product of the portion of the area of the projection surface occupied by the sprite and a measure of rendering error that is computed as the sum of squared distance between a set of reference points in the warped and a perfectly rendered sprite. De-tails about the use of characteristic points to develop a fiducial that reports a quality of approximation are described by Lengyel and Snyder in [19].

We now need to extend the cost for individual sprites into models of cost for multiple sprites. A simple model for combining the cost assigned to multiple sprites into a total perceptual cost $C^p$ is the sum of all of the costs associated with each sprite,

$$\mathcal{C}^p = \sum_i C^p(R_k, S_i) \tag{1}$$

where $R_k$ is instantiated to the rendering action taken for each sprite.

The combination of the cost of sprites need not be additive as described in Equation 1. For example, the perceived losses in the quality of an image may be a more complex function of the degradation of multiple sprites, dependent on the relationships among sprites and the way sprites comprise objects. A number of phenomena could arise from perceptual dependencies among degradations of multiple sprites, including amplification and pop-out effects, and perceptual phenomena stemming from degradations of sets of sprites perceived to comprise the same object versus sprites scattered among different objects. For example, we can employ more complex functions for combining the costs of the degradations of multiple sprites within single objects.

## 5.2 Visual Attention and Expected Cost

Findings in visual perception make us keenly interested in the influence of a viewer's focus of attention and of the overall scene complexity on a viewer's perception of the degradations associated with rendering approximations. As highlighted by the set of schematized curves in Figure 5, we seek to understand the influence of attention on the functions linking perceptual quality to various approximations of rendering sprites or larger objects. After describing models of attention, we shall mesh the models of attention and models of cost described above to generate a model of *expected cost* of image degradation.

As we reviewed in Section 3, key issues about attention are unresolved in the psychological literature. Nevertheless, we can build expressive models with the ability to capture the expectation that the cost of rendering approximations will diminish with diminishing attention to sprites.

Let us extend the perceptual cost function by taking as an additional input a variable $A$ capturing the attentional focus on a sprite, $C^p(R_k, S_i, A)$. The attentional focus can be considered to be a scalar measure of the degree to which visual resources are allocated to portions of a screen, or, alternatively, as a discrete, binary variable that models a viewer as either attending or as not attending to specific elements of an image. We seek to compute an expected cost of a rendered scene by approximating the probability, $p(A^{S_i}|E)$, that a user is selectively focusing on sprite $S_i$ given some evidence, $E$. Evidence can include information about the structure of a scene, and, where relevant, about the goals associated with an interactive task such as a computer game.

In a *continuous attentional model*, we assume a scalar measure of attention as a random variable that varies between representing a minimal amount of attention at zero and a maximal amount of attention at one. With this model, the expected perceptual cost is,

$$\mathcal{EC}^p = \sum_i \int_{x=0}^1 p(A^{S_i} = x|E)C^p(R_k, S_i, x)dx \quad (2)$$

We can simplify this model by factoring the attention variable out of the cost function. In such a model, we revert to the attention-independent form of cost function which represents the quality of a rendered sprite when attention is fully focused on that sprite. We diminish the cost by a multiplicative attention factor, $\alpha(x)$, which ranges between zero and one, as a viewer's attention varies between no attention and full attention to a sprite,

$$\mathcal{EC}^p = \sum_i \int_{x=0}^1 p(A^{S_i} = x|E)\alpha(x)C^p(R_k, S_i)dx \quad (3)$$

We can simplify Equation 3 to the *binary attentional model* case where we consider the likelihood that a user either selectively attends to an object or does not attend to the object, and that the viewer perceives the full cost of a degradation when attending and a diminishment of the cost when not attending,

$$\mathcal{EC}^p = \sum_i p(A^{S_i}|E)C^p(R_k, S_i)$$
$$+[1 - p(A^{S_i}|E)]\alpha C^p(R_k, S_i) \quad (4)$$

where $\alpha$ is a constant factor.

We can further simplify the binary model by assuming that $\alpha$ is zero, implying that sprites that are not receiving attention do not contribute to the cost of scene,

$$\mathcal{EC}^p = \sum_i p(A^{S_i}|E)C^p(R_k, S_i) \quad (5)$$

### 5.3 Conditioning on Contiguous Objects

Although we can cast all of our equations for attention and expected cost in terms of individual sprites, it can be useful to jump to a model of attention based on the natural tendency for viewers to attend to *contiguous objects*. That is, we consider the probability,

$p(A^{S_{ij}}|A^{O_j}, E)$, of attending to an image element conditioned on the viewer selectively attending to a set of interrelated elements that are perceived as contiguous object, $O_j$. For example, if we substitute into the discrete attention model (Equation 4) the probabilities of attending to objects and the conditional probabilities of attending to sprites given the focus on the specific objects, the expected cost is

$$\mathcal{EC}^p = \sum_i \sum_j p(A^{S_{ij}}|A^{O_j}, E)p(A^{O_j}|E)C^p(R_k, S_{ij})$$
$$+[1 - p(A^{S_{ij}}|A^{O_j}, E)p(A^{O_j}|E)]\alpha C^p(R_k, S_{ij}) \quad (6)$$

where $p(A^{O_j}|E)$ is the probability of a user attending to an object, and $p(A^{S_{ij}}|A^{O_j}, E)$ is the probability that a user will attend to elements $ij$ of object $j$, given that the viewer's attention is drawn to that object, and $C(R_k, S_{ij})$ is the perceptual cost of applying degradation strategy $R_k$ to render element $S_{ij}$. For the simpler binary model, the expected cost is just,

$$\mathcal{EC}^p = \sum_i \sum_j p(A^{S_{ij}}|A^{O_j}, E)p(A^{O_j}|E)C^p(R_k, S_{ij}) \quad (7)$$

The models of expected cost of rendered scenes provide a framework for studying the perception of image quality as a function attention and of multiple degradations. They also provide an outline for constructing and using models of cost and attention to control rendering.

We can build upon these basic models to develop the means for representing such potential attentional mechanisms as the tradeoff between the radial breadth of attention and the degree of attention (the *zoom-lens model*), by introducing additional structure such as parameters that describe the probability distribution over the degree of attention as a function of the radial distance from objects at the center of attention.

## 6　Models of Attention

In the context of our models, the more we know about the selective attention of viewers to particular aspects of a scene, and the smaller that $\alpha$ is, the more we can enhance the perceived quality of images under computational resource constraints by selectively degrading sprites that have minimal expected cost. We seek to characterize the time-dependent probability distribution over a viewer's attention as a function of distinctions in images being viewed, the task at hand, and context. Probabilistic models of a viewer's attention can range from coarse hand-authored approximations based on heuristics about attention to more complex models learned from data about the gaze of viewers, conditioned on content and task.

Some computer applications provide a set of task-oriented contexts that can provide rich inputs to prob-abilistic models of attention. For example, the time-varying goals and point systems that are defined in computer games can be used to generate probability distributions over the attention being directed at graphical objects. In, gaze-tracking studies of user's playing computer arcade games at the Microsoft usability labs, we have found that game contexts are induce prototypical patterns of gaze in relation to rendered objects.

Beyond the task-oriented models, we have been working to characterize attention as a function of distinctions that characterize graphics content. Some evidence about objects can be gleaned directly from graphics content. For example, graphics systems have access to such information as the number of objects, as well as the size and relationships among objects in a scene. Information about the visual angle subtended by objects and the virtual distance from objects to the viewers can serve as inputs to models of attention.

Beyond automated approaches, we can rely on the expert judgments of the authors of graphics content. We are working to provide a language and associated authoring tools that allow authors to specify information about the likely pattern of a viewer's attention for scenes in rendered movies. The goal is to ease the burden on authors by providing stereotypical models of attention that take as inputs an author's high-level descriptions about the viewer's attention. Rather than require authors to provide inputs about the detailed priorities of multiple sprites, we allow authors to specify *primary foci* or to classify objects in scenes as elements of one of several key groups of objects such as *primary actors, secondary actors, critical environment,* and *background environment*. Such foci or groups serve as arguments to models of attention that generate the likelihoods that other objects in a scene will be noticed by a viewer *given* these foci and additional evidence about the scene. Evidence includes features from the scene that capture significant spatial and temporal relationships with objects at the focus of attention such as the class of object, size of object, and optical distance of the object from the viewer. We can employ spotlight or zoom lens model and make the size of the scope of attention a constant or a function of properties of the objects in an image.

At run-time, the likelihoods that a user will attend to objects in each class of object, $p(G)$ is assigned and the probabilities of the $n$ objects in each of the groups is assigned a probability modulated by key details about the configuration of the objects. Once objects have been assigned probabilities, conditional probabilities of attending to the sprites comprising the objects are assigned. We take as inputs the key foci of attention, $A^{O^*}$, and employ functions of features in the scene to approximate the conditional probabilities, $p(A^{O_i}|A^{O^*}, E)$. These conditional probabilities can be substituted for $p(A^{O_i}|E)$ in the expected cost models.

The probabilities that a viewer will attend to a specific sprite comprising an object, $p(A^{S_{ij}}|A^{O_j}, E)$, can be generated as a function of such factors as whether the sprite defines an edge of an object, the portion of the surface area of the total object occupied by sprites, and the degree to which the approximately rendered sprite does not naturally fit into the object. The latter can be captured as a function of the perceptual cost described earlier. That is, we condition the attention partly on the cost, $p(A^{S_{ij}}|A^{O_j}, E, C^p)$. Here the perceptual cost provides input to the *probability* that the user will attend to the sprite—in addition to serving as the *cost* component of the expected cost model.

There is opportunity for learning probability distributions over attention, conditioned on information about a scene, including an author's input. Such models would allow for Bayesian inference to diagnose attention. Inference in real-time would be restricted to the solution of tractable inference models (e.g., naive Bayesian diagnosis). However, more complex inferential models may be of value in the offline assignment of attention for use in real-time rendering.

### 6.1 Consideration of Cost over Multiple Frames

Although we have focused on cost associated with rendering sprites for single frames, it can be important to consider the cost over a set of recent frames. Such richer cost functions must consider the contributions of the visual *persistence* of an artifact across frames. Consideration of the expected cost over the last $n$ frames, $C^p(t-n)$, requires modeling the cost and attention as a function of the persistence and dynamics over multiple frames. A viewer is more likely to notice an artifact in an object that persists over time. Thus, we must consider in parallel the influence of the persistence of artifacts on the probabilities that a user will attend to sprites, conditioned on the recent history of error, $p(A^{S_{ij}}|A^{O_j}, E, C^p(t-n))$. Procedurally, for each sprite, we must maintain a list of recent error associated with that sprite in individual adjacent or *recent* frames and combine this list into an overall cost to evaluate the current probabilities and costs.

## 7  Rendering Regulation Policies

Models of the expected utility of scenes or sequences of scenes provide critical metrics for guiding search within a large space of rendering policies. Our goal is to create images of the highest quality by minimizing the expected perceptual cost of frames, subject to the constraints on available resources. For graphics architectures like Talisman that commit to a constant frame rate, the time for completely rendering a scene may be less than the time available. In systems that allow for the variation of frame rate, we must con-sider the cost of diminishing the frame rate along with the other kinds of degradations that can be performed. Clearly, a general search involving the consideration of all feasible degradation actions for each component of a scene is intractable. However, we can employ approximations that take advantage of measures of expected utility and that consider in order key classes of approximation.

Let us focus on the control of regulation in the Talisman architecture. The two key dimensions of rendering approximation in the Talisman architecture are temporal and spatial resolution. Decisions can be made about the spatial resolution of sprites and about the error tolerated in the reuse of individual sprites through inexpensive warping procedures before they are re-rendered with costly procedures. We seek to render images of the highest quality by choosing to render a set of sprites and applying inexpensive two-dimensional transformations to the remaining sprites such that the expected perceptual cost of a frame or a set of frames is minimized subject to the constraint that the total time is less than or equal to the time available for generating the frame, given the target frame rate.

The cost of postponing the re-rendering of a sprite and using instead a two-dimensional transformation of a sprite that was rendered earlier is the degradation of the overall *perceived* quality of frames due to that sprite. The benefits are the computational savings incurred by putting off the rendering and relying instead on a two-dimensional transformation of an earlier sprite. Warping can impose geometric and temporal artifacts that become more salient as sprites that have been warped for several frames are replaced with a re-rendered sprite. By increasing the tolerated error, we save on resources by minimizing re-rendering, but may introduce noticeable artifacts in object distortion, problems with visibility of overlapping sprites, and discontinuities.

We have built regulation policies for Talisman that consider in sequence policies for re-rendering versus warping sprites followed by spatial degradation. The methods center on a consideration of the marginal costs and benefits of the degradations.

The *marginal computational cost*, $\Delta C^c(R_r, R_w, S_i)$ of re-rendering versus warping a sprite is the difference in computational resources required by these two rendering actions,

$$\Delta C^c(R_r, R_w, S_i) = C^c(R_r, S_i) - C^c(R_w, S_i) \qquad (8)$$

where $C^c(R_k, S_i)$ is the computational cost of taking rendering action $R_k$ for sprite $S_i$. The perceptual models gives the system access to the marginal computational benefit and cost for each sprite.

From Equation 3, we see that the perceptual cost contributed to the overall expected perceptual cost of the whole image by reusing each sprite through a two-dimensional warp (*i.e.*, rendering action $R_w$) is,

$$\Delta C^p(R_r, R_w, S_i) = \int_0^1 p(A^{S_i} = x|E)\alpha(x)C^p(R_w, S_i)dx \qquad (9)$$

The maximization of image quality for dimension of temporal resolution under the deadline defined by the Talisman frame rate maps to the knapsack problem [10]; we seek to maximize the expected value of items within a "knapsack" sized by computational resources allowed by the system's frame rate. Although the knapsack problem is NP-Complete, we can use information about the marginal costs and benefits of rendering sprites in a useful approximation.

Rerendering each sprite reduces the overall perceptual cost by this incremental amount in return for the marginal computation required for rendering the sprite. The ratio of the incremental gain in quality and the computational cost, gives us a measure the expected *perceptual refinement rate* with computation, $\Phi(S_i)$,

$$\Phi(S_i) = \frac{\Delta C^p(R_r, R_w, S_i)}{\Delta C^c(R_r, R_w, S_i)} \qquad (10)$$

We can employ a greedy algorithm to minimize the expected cost of a rendered scene by taking advantage of the $\Phi(S_i)$ associated with sprites. We order sprites for re-rendering by $\Phi(S_i)$ until reaching the computational deadline and compare the perceptual cost of this policy with the policy of rendering only the sprite with the highest marginal value. Choosing the policy of these two with the minimal perceptual cost can be shown to be of value within a factor of two of the minimal cost policy [10]. For some additional control cost, we can employ a knapsack approximation procedure involving limited search among subsets of sprites to bring the perceptual cost even closer to the optimal value [26].

## 7.1    Greedy Optimization for Multiple Dimensions of Approximation

Moving beyond decisions within the realm of temporal resolution, we can consider other dimensions of degradation by taking advantage of the knapsack approximation for a means of efficiently comparing policies. Although general search through the space of changes for other dimensions, such as spatial degradation of sprites, is intractable, we can quickly compare policies after making greedy interventions along other dimensions of perceptual loss. After modification of the $\Phi(S_i)$ associated with sprites following consideration of additional degradations, we re-evaluate the expected cost of the rendering plan generated by the knapsack approximation.

In one myopic approach, we explore in order of predefined priority, diminishing the texture level-of-detail,

then the geometric level-of-detail, then spatial sampling, and finally shading complexity. For each class of degradation, we consider the effects of making a predefined amount of degradation in return for the gain associated with the ability to re-render additional sprites with resources made available by the degradation. We can prune the consideration of sprites by examining the best increase in overall image quality that is possible with the rendering of one additional sprite–the sprite that is the first pruned from the rendering list because of the deadline. If no myopic changes at one dimension will reduce the expected cost of the scene, we move onto the next dimension. After all dimensions of degradation are considered we execute the policy.

### 7.2 Investigation of Regulation Policies

We have experimented with the use of expected cost modeling and regulation with a simulation of Talisman and have generated scenes demonstrating the value of the control policies. For our studies, we explored several perceptual cost models. In one set of studies, we used as the perceptual cost the product of the screen area occupied by a sprite multiplied by a linear combination of our fiducial measure of geometric distortion and degradation of resolution of that sprite, informed by the results of perceptual studies. We tuned constants to weight the contribution of the two dimensions to the overall perceptual cost of a sprite. The Talisman architecture provides us with an estimate of the cost of rendering each sprite as a function of the rendering actions.

To construct a model of attention, we tagged objects by hand within individual scenes as being in the group of *primary actors*, *secondary actors*, *critical environment*, and *background environment*. The expected cost model described in Equation 4 was employed for several values of $\alpha$, capturing the degree to which the degradation of sprites of objects that are not at the center of attention are noticed and, thus, add to the overall perceptual cost. As $\alpha$ is moved to zero, the system assumes that rendering approximations that are not being attended to can be better tolerated and selectively degrades the sprites composing those objects.

We are continuing our perception studies to ascertain details about the relationships among a variety of approximations available in Talisman and perceptions of image quality. We are also continuing studies on details of the influence of attention on the perception of quality. Results of the perceptual and attentional studies with human subjects will be valuable to further refine the models for computing the expected cost of images and for maximizing the quality of rendered images.

## 8 Future Work and Summary

We have described key issues with the regulation of graphics approximation procedures based on an expected-cost approach. We reviewed central findings on visual search and attention from the Cognitive Psychology literature and presented expected-cost models that can take advantage of information about the perceptual costs associated with rendering approximations. We described our use of the models to generate rendering regulation policies. The approach has been validated with sample rendering sequences. We have found value in assessing and manipulating separately the key components of attention and perceptual cost to maximize the perceptual quality of images under limited resources.

We are studying several approaches to the control of rendering that build upon the basic control strategies we have described. We also seek to understand the benefits of moving beyond additive models of the cost of multiple degradations to consider interdependencies associated with rendering error in related sprites. For example, we are interested in the perceptual costs associated with heterogeneities in the quality at which interrelated sprites or objects are rendered over time and space.

We are continuing our research on assessing and refining models of the expected computational and perceptual costs of rendering approximations. Enhancing the fidelity of these models will strengthen the ability of graphics systems to selectively degrade components of rendered images in a manner that directs artifacts of approximation into the more tolerant "blind spots" of the human visual system.

### Acknowledgments

We thank Mary Czerwinksi and Kevin Larson for providing insights and pointers on psychological studies of visual search and for assistance with experiments on visual perception and attention, Jim Kajiya for sharing his insights about tradeoffs in graphics rendering, John Snyder for input on regulation and layered architectures, Hugues Hoppe for input on model simplification strategies, Bobby Bodenheimer and Larry Ockene for their input on the details of the Talisman architecture, and Mike Jones for ongoing discussion on links to work on resource-aware operating systems.

### References

[1] J. Beck and B. Ambler. The effects of concentrated and distributed attention on peripheral acuity. *Perception and Psychophysics*, 14:225–230, 1973.

[2] P. Cavanagh, M. Arguin, and A. Triesman. Effects of surface medium on visual search for orientation and size features. *Journal of Experimental*

*Psychology: Human perception and Performance*, 16:479–491, 1990.

[3] S.E. Chen and L. Williams. View interpolation for image synthesis. In *Proceedings of SIGGRAPH 93*, pages 279–288. SIGGRAPH, August 1993.

[4] R. Colegate, J.E. Hoffman, and C.W. Eriksen. Selective encoding from multielement visual displays. *Perception and Psychophysics*, 14:217–224, 1973.

[5] H. Egeth. Attention and preattention. In G.H. Bower, editor, *The Psychology of Learning and Motivation*, volume 11, pages 277–320. 1977.

[6] C.W. Eriksen and J.D. St. James. Visual attention within and around the field of focal attention: A zoom lens model. *Perception and Psychophysics*, 40:225–240, 1986.

[7] C.W. Eriksen and Y. Yeh. Allocation of attention in the visual field. *Journal of Experimental Psychology: Human Perception and Performance*, 11:583–597, 1985.

[8] C.W.W. Eriksen and J.E. Hoffman. Temporal and spatial characteristics of selective encoding from visual displays. *Perception and Psychophysics*, 12:201–204, 1972.

[9] T.A. Funkhouser and C.H. Sequin. Adaptive display algorithm for interactive frame rates during visualization of complex virtual environments. In *Proceedings of SIGGRAPH 93*, pages 247–254. SIGGRAPH, August 1993.

[10] M.R. Garey and D.S. Johnson. *Computers and Intractability: A Guide to the Theory of NP-Completeness*. W.H. Freeman and Company, New York, 1979.

[11] P. Goolkasian. Retinal location and its effect on the processing of target and distractor information. *Journal of Experimental Psychology: Human Perception and Performance*, 7:1247–1257, 1981.

[12] H. Hoppe. Progressive meshes. In *Proceedings of SIGGRAPH 96*, pages 99–108. SIGGRAPH, August 1996.

[13] E. Horvitz and J. Lengyel. Flexible rendering of 3D graphics under varying resources. In *Fall Symposium on Flexible Computation, Cambridge MA, Report FS-96-06*, pages 81–88. AAAI: Menlo Park, CA, November 1996. Also available as Microsoft Research Technical Report MSR-TR-96-18, Winter 1996.

[14] E. Horvitz, J. Lengyel, K. Larson, and M. Czerwinski. Harnessing perception of image quality to guide graphics rendering. Technical Report Microsoft Research Technical Report MSR-TR-97-12, Microsoft Research, Spring 1997.

[15] G.W. Humphreys. On varying the span of visual attention: Evidence for two modes of spatial attention. *Quarterly Journal of Experimental Psychology*, 33A:17–31, 1981.

[16] J. Jonides. Further toward a model of the mind's eye's movement. *Bulletin of the Psychonomic Society*, 21:247–250, 1983.

[17] B. Julesz. A theory of preattentive texture discrimination based on first order statistics of textons. *Biological Cybernetics*, 41:131–138, 1975.

[18] R.A. Kinchla. Detecting target elements in multielement arrays: A confusability model. *Perception and Psychophysics*, 15:149–158, 1974.

[19] J. Lengyel and J. Snyder. Rendering in a layered graphics architecture. In *Proceedings of SIGGRAPH 97*. SIGGRAPH, August 1997.

[20] L. McMillan and G. Bishop. Plenoptic modeling: an image-based rendering system. In *Proceedings of SIGGRAPH 95*, pages 39–46. SIGGRAPH, August 1995.

[21] U. Neisser. Decision time without reaction time: Experiments in visual scanning. *American Journal of Psychology*, 76:376–383, 1980.

[22] P. Podgorny and R. Shepard. Distribution of visual attention over space. *Journal of Experimental Psychology: Human Perception and Performance*, 9:380–394, 1983.

[23] M. Posner. Orienting of attention. *Quarterly Journal of Experimental Psychology*, 32:3–27, 1980.

[24] W. Prinzmetal and W.P. Banks. Perceptual capacity limits in visual detection and search. *Bulletin of the Psychonomic Society*, 21:263–266, 1983.

[25] M. Regan and R. Pose. Priority rendering with a virtual reality address recalculation pipeline. In *Proceedings of SIGGRAPH 94*, pages 155–162. SIGGRAPH, August 1994.

[26] S. Sahni. Approximate algorithms for the 0/1 knapsack problme. *J. ACM*, pages 115–124, 1975.

[27] J. Shade, D. Lischinski, D. H. Salesin, T. DeRose, and J. Snyder. Hierarchical image caching for accelerated walkthroughs of complex environments. In *Proceedings of SIGGRAPH 96*, pages 75–82. SIGGRAPH, August 1996.

[28] M.L. Shaw. A capacity allocation model for re-action time. *Journal of Experimental Psychology: Human Perception and Performance*, 4:586–598, 1978.

[29] R.M. Shiffrin and G.T. Gardner. Visual process-ing capacity and attentional control. *Journal of Experimental Psychology*, 93:72–82, 1972.

[30] G.L. Shulman, J. Wilson, and J.B. Sheehy. Spa-tial determinants of the distribution of atten-tion. *Perception and Psychophysics*, 37(16):59–65, 1985.

[31] G. Sperling and M. J. Melchner. The attention operating characteristic: Examples from visual search. *Science*, 202:315–318, 1978.

[32] J. Torborg and J. Kajiya. Talisman: Commodity realtime 3D graphics for the pc. In *Proceedings of SIGGRAPH 96*, pages 353–363. SIGGRAPH, August 1996.

[33] A.M. Triesman and G. Gelade. A feature-integration theory of attention. *Cognitive Psy-chology*, 12:97–136, 1980.

[34] J.M. Wolfe, K.R. Cave, and S.L. Franzel. Guided search: An alternative to the feature integration model for visual search. *Journal of Experimental Psychology*, 15(3):419–433, 1989.